\documentclass[10pt,twocolumn,letterpaper]{article}

\usepackage{wacv}
\usepackage{times}
\usepackage{epsfig}
\usepackage{graphicx}
\usepackage{amsmath}
\usepackage{amssymb}

\wacvfinalcopy 

\def\wacvPaperID{618} 

\ifwacvfinal\fi
\begin{document}


\title{GAN-based Pose-aware Regulation for Video-based Person Re-identification}


\author{
\vspace{0.5em}
  Alessandro Borgia\textsuperscript{1,2}, Yang Hua\textsuperscript{3}, 
  Elyor Kodirov\textsuperscript{4}, Neil M. Robertson\textsuperscript{3}\\
  \vspace{0.5em}
  \textsuperscript{1}Heriot-Watt University,  \textsuperscript{2}University of Edinburgh, \textsuperscript{3}Queen's University Belfast, \textsuperscript{4}Anyvision\\
  \texttt{ab41@hw.ac.uk,\{y.hua,n.robertson\}@qub.ac.uk, elyor@anyvision.co}
}

\maketitle
\ifwacvfinal\thispagestyle{empty}\fi

\begin{abstract}
Video-based person re-identification deals with the inherent difficulty of matching unregulated sequences with different length and with incomplete target pose/viewpoint structure. Common approaches operate either by reducing the problem to the still images case, facing a significant information loss, or by exploiting inter-sequence temporal dependencies as in Siamese Recurrent Neural Networks or in gait analysis. However, in all cases, the inter-sequences pose/viewpoint misalignment is not considered, and the existing spatial approaches are mostly limited to the still images context. To this end, we propose a novel approach that can exploit more effectively the rich video information, by accounting for the role that the changing pose/viewpoint factor plays in the sequences matching process. Specifically, our approach consists of two components. The first one attempts to complement the original pose-incomplete information carried by the sequences with synthetic GAN-generated images, and fuse their feature vectors into a more discriminative viewpoint-insensitive embedding, namely Weighted Fusion (WF). Another one performs an explicit pose-based alignment of sequence pairs to promote coherent feature matching, namely Weighted-Pose Regulation (WPR). Extensive experiments on two large video-based benchmark datasets show that our approach outperforms considerably existing methods. 

\end{abstract}

\begin{figure*}
	\begin{center}
		\includegraphics[width=150mm]{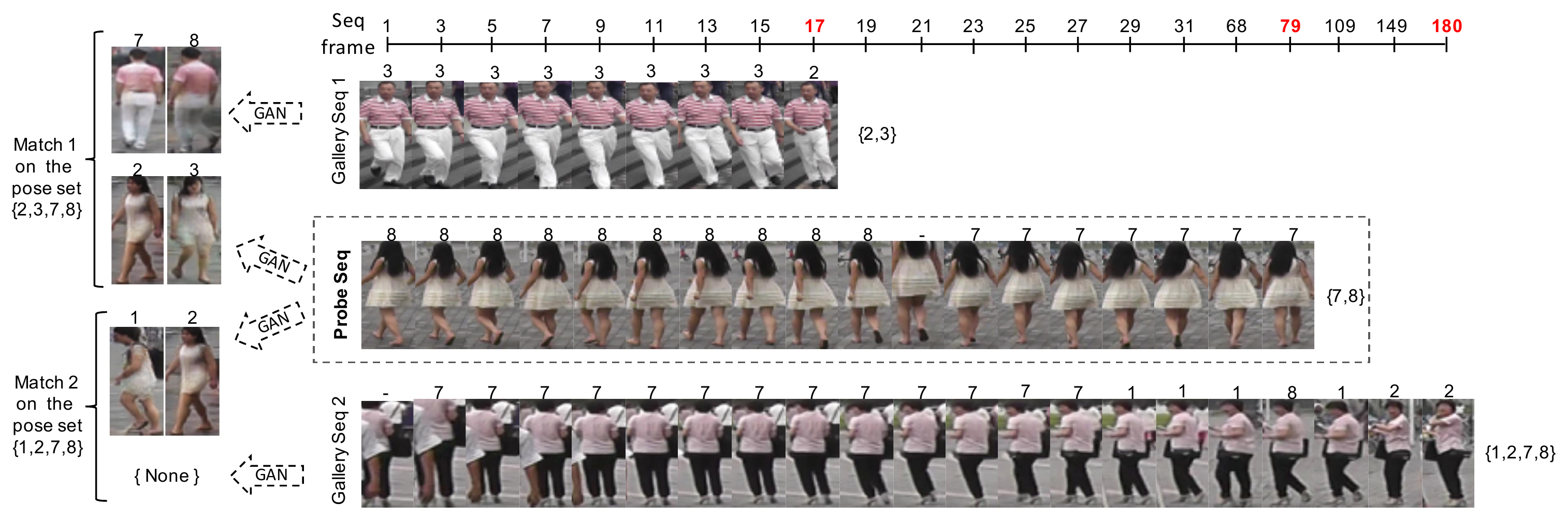}
        \end{center}
		\caption{MARS \cite{zheng2016mars} sequences with different length and unregulated pose/viewpoint at corresponding time frames. The probe sequence (in the middle) has to be matched against all the sequences of the gallery (top and bottom). For each (probe, test) sequence pair, the missing poses are produced and the corresponding images are generated by a GAN and added to the original sequences. For the sake of space constraints, the frame sequence axis is not linear. From the top to the bottom: IDs$=\{23, 1, 967\}$. Best viewed in color.}
		\label{fig:3seq}
\end{figure*}

\section{Introduction}

In surveillance, person re-identification (re-id) has emerged as a fundamental capability that no tracking system aiming to operate over a wide area network of disjoint cameras can, concretely, renounce to have. 
Recently, the person re-id task has broadened from the still images context \cite{wang2016joint, varior2016siamese, wu2016enhanced, borgia2018cross, ahmed2015improved} to video-based approaches, either supervised \cite{wu2016deep,mclaughlin2016recurrent,zheng2016mars,you2016top,wang2016person,huang2018video} or semi-unsupervised \cite{ye2017dynamic,liu2017stepwise,wu2018exploit}. For the video-based person re-id, given a query sequence of images (a.k.a., tracklet) of the target of interest, the challenge consists of identifying all the corresponding matching tracklets captured across the network. Dealing with full sequences of images as opposed to single images offers significant advantages: a) exploiting the temporal dependencies between \textit{intra}-sequence frames \cite{wu2016deep,mclaughlin2016recurrent,xu2017jointly,zhou2017see}; b) extracting more robust spatial appearance descriptors \cite{zheng2016mars, wojke2018deep,you2016top}; c) partially recovering from occlusions and reducing the influence of the background \cite{wang2014person,wang2016person,huang2018video}. All these three aspects contribute to reducing the impact of the factors affecting performance (changing pose/viewpoint, lighting, occlusions, etc.), due to the availability of more diverse samples of the same identity. Depending on whether the frame-level feature aggregation scheme is based on temporal cues, estimated noise-state of frames, motion-state or pose/viewpoint (that is spatial cues), we can identify different \textit{alignment} methods for matching corresponding video-fragments features (see Sec.\ref{alignment_methods} for related references).

Temporal alignment methods usually suffer from noisy unregulated sequences with severe viewpoint and lighting variability \cite{wang2016person}. Under such scenarios, it is difficult to estimate accurately gait phase and cycle or the optical flow energy profile as in \cite{zhang2017learning}. Furthermore, the background clutter and occlusions from other people (in crowded context) interfere with the target feature map computation, causing over-fitting. Differently, other approaches, disregarding inter-frames temporal dependencies, focus on building more robust spatial features by performing an overall average pooling operation across the entire feature maps sequence \cite{zheng2016mars, wojke2018deep, zhong2017identification, you2016top}, thus reducing the multiple feature maps to a single instance. Despite performance benefit from it, we argue that this summarizing operation is not performed optimally due to the following reasons: a)  it is frame-indiscriminate and ignores the pose/viewpoint information; b) real sequences are typically pose-incomplete, therefore the extracted features may not be as discriminative as under the availability of complete pose-information. A few techniques targeting occlusion impact reduction divide the incomplete unaligned sequences according to the estimated noise-state of the frames \cite{huang2018video} or their motion-state (from the optic flow intensity profile) \cite{wang2014person,wang2016person}, or weight samples by a quality-aware network \cite{liu2017quality}.

Our work originates from the observation that none of the mentioned approaches aiming to exploit the extra information of video-data account for the relevance of the viewpoint information in the sequence pairs matching process, despite the pose/viewpoint problem has been largely investigated in the image-based re-id  \cite{zheng2017pose, zheng2017pedestrian, qian2017pose, ma2017pose, sarfraz2017pose}. We embrace the view that the need of accounting for pose/viewpoint is even greater in video-based re-id than in still images-based re-id. As proved in \cite{you2016top,zheng2016mars}, the ambiguity in distinguishing between video-based person representations can increase with respect to image-based representations, because the extra information associated with the motion is hardly discriminative. In order to incorporate the pose/viewpoint information in the sequences matching process for video-based person re-id, we propose a Pose-Driven Sequence Regulation (\textbf{PDSR}) approach articulated in two complimentary sub-schemes, the Weighted Fusion (WF) scheme and the Weighted Pose-Regulation (WPR) scheme. The first one enriches each sequence with more varied and complete viewpoint information by adding synthetic images, which are generated by a Generative Adversarial Network (GAN) with conveniently defined canonical poses (similarly to \cite{qian2017pose} for still images). This helps to synthesize people representations which generalize better to unknown identities, as shown in Figure \ref{fig:3seq}. The second one (WPR) explicitly aligns sequences pairs by canonical poses to match spatial information coherently. Both schemes, differently from \cite{liu2018pose}, operate at testing time, conceptually similar to the viewpoint repositioning strategy in \cite{savarese2008view} at recognition time. In summary, the main contributions of this paper are the following:
\begin{itemize}
    \item[--] To the best of our knowledge, this is the first work to apply a GAN-based generative model to video-based person re-identification for complementing and pose-aligning the original incomplete data.
    \item[--] We identify the importance of pose-based coherent matching of sequences to exploit more effectively the available video-information and synthesize feature vectors with improved generalization capability.
    \item[--] Our approach achieves a significant performance boost on two large video-based person re-id benchmarks with comparison to recently proposed techniques.
\end{itemize}

\section{Related Work}

\subsection{Cross-View Invariant Techniques}
The importance of accounting for the pose/viewpoint invariance problem, in person re-id, has been amply proved by many works. A popular approach is metric learning \cite{zhang2017learning,you2016top,zhu2016videobased,li2013locally,wojke2018deep, wang2019deep} where a similarity metric is learned in the space of the video-level feature vectors expressing different views, aiming to increase the intra-class compactness and the inter-class distance of the identities. 
A different stream of research, complementary to metric learning, tackles the viewpoint problem by focusing on designing/learning more robust feature representations, for example, exploiting the temporal aggregation of multiple frame-level features maps \cite{mclaughlin2016recurrent} or performing spatial fusion/concatenation of global/local features \cite{xu2017jointly,chen2017deep}.

The introduction of GANs has represented a step forward in the way that the viewpoint problem is tackled because it allows creating new synthetic data under the desired viewpoints without requiring any extra labelling \cite{qian2017pose,ma2017pose,siarohin2017deformable,zhang2018crossing,zhao2017multi}. This is why it lends itself particularly well to be combined with the existing video-based feature extraction techniques for producing even more discriminative embeddings. In this paper, we follow this combined approach for complementing the incomplete pose information of video-sequences and refer to \cite{siarohin2017deformable} for the GAN architecture and to \cite{wojke2018deep} for the feature extraction CNN.


\subsection{Video-Sequences Alignment Techniques}
\label{alignment_methods}
Video-based person re-id demands to deal with the \textit{intra}-sequence appearance variability caused by many changing factors across cameras (occlusions, pose and viewpoint, etc.), in addition to handling the visual ambiguity at the \textit{inter}-matching items level as it happens in the still images case. To tackle this, a stream of literature has emerged proposing temporal \cite{simonnet2012re,gao2016temporally,mclaughlin2016recurrent,xu2017jointly}, spatio-temporal \cite{liu2015spatio,ma2017person}, pose/viewpoint-based \cite{suh2018part,zhao2017deeply,wang2018transferable,qian2017pose,wei2017glad}, motion-state-based \cite{wang2014person,wang2016person} and noise-state-based \cite{huang2018video} sequence alignment techniques. Gao {\it et al.} \cite{gao2016temporally} proposes a temporally aligned pooling representation for video-based person re-id which relies on dividing a sequence into several segments according to the sinusoid of a person walking cycle and then performing segment-based pooling to extract a representation. The periodicity of pedestrians gate is exploited also in \cite{liu2015spatio} to generate a spatio-temporal body-action model made up of a series of action primitives of certain body parts treated independently from each other. The strict alignment assumptions made by gait recognition techniques are relaxed in \cite{wang2016person,zhang2017learning} where unregulated video-sequences are automatically broken down based on motion energy profiling (e.g., optical flow). With the spreading of the deep learning paradigm, temporal-based approaches have emerged using RNNs (combined with Convolutional Neural Networks) in Siamese configuration \cite{mclaughlin2016recurrent,xu2017jointly,wu2016deep,zhou2017see}. These methods embed both the mid-term temporal information in the video-sequence representation and the long-term appearance information summarized by the temporal pooling layer. Another form of sequence alignment is the spatial regulation, handling the body part misalignment problem across frames, usually addressed by defining part-based body models, either pre-defined (stripes or grids) \cite{cheng2016person,wu2016enhanced,varior2016siamese} or learned from data by spatio-temporal attention mechanisms \cite{suh2018part,zhao2017deeply,xu2018attention,li2018diversity} that take into account the non-rigid shape of the human body.

\subsection{GANs for Person Image Generation}
Generative Adversarial Networks (GANs) \cite{goodfellow2014generative} represent the most popular approach to deep learning-based generative image modelling, compared to either the unsupervised techniques based on variational autoencoders \cite{kingma2013auto} or autoregressive models \cite{oord2016pixel}. A distinction in two classes of GANs can be made depending on the nature of the input distribution to the generator:
noise-driven GANs \cite{zheng2017unlabeled,ma2018disentangled} where a mapping from a Gaussian noise distribution to the images distribution is learned and attribute conditional GANs \cite{isola2017image}. As to the former class, \cite{ma2018disentangled} represents for the person re-id task a notable work where a disentangled person representation with respect to pose, background and foreground is learned. Zheng {\it et al.} \cite{zheng2017unlabeled} applies DCGAN \cite{radford2015unsupervised} to the person re-id task to improve the discriminative learning by assigning a uniform label distribution of the generated images over all the existing classes. Therefore, it assumes that the generated data belong to none of the training classes, differently from \cite{salimans2016improved} where the synthetic new samples are treated as a single extra class. In our work, instead, following \cite{qian2017pose}, we assign identity membership labels to the generated images in viewpoint-based canonical forms, leveraging the good quality of the synthetic data, instead of using them as a regularization strategy for outliers. The noise-based GANs formulation is less effective in person re-id due to the challenging nature of the task that requires to condition the generated images to some attributes like pose \cite{ma2017pose,qian2017pose,zhao2017multi, siarohin2017deformable}, clothing \cite{lassner2017generative}, camera style \cite{zhong2018camera}, dataset style \cite{wei2017person, deng2018image} for domain adaptation. We adopt the conditional architecture proposed in \cite{siarohin2017deformable} because of its capability to handle the global morphological transformation happening between the input-target poses pair.

\section{Proposed Method}
\label{sec:prop}

Our approach consists of considering the pose/viewpoint role, in terms of both completeness and alignment. In order to match together a pair of sequences, we convert each of them into a viewpoint-normalized form made up of a set of pre-defined canonical poses manually selected from the train set. As to the pose-information completeness aspect, the basic idea of our technique is that before matching two sequences they should be enriched with complete appearance information of the observed target. In other words, they should contain instances of the entire set of possible discrete viewpoints in which the pedestrian may be shot. Integrating this information in all sequences allows extracting better informative deep representations of the identities which in turn translates into a more successful features matching. The viewpoint instances of a target identity, which are not present in the original data, need to be generated. This is done by conditional GANs that can generate synthetic images of pedestrians in any desired pose \cite{siarohin2017deformable, qian2017pose, zheng2017unlabeled, zhang2018crossing, walker2017pose, ma2018disentangled, huang2018multi, zhao2017multi, ma2017pose} and dataset style \cite{wei2017person, jetchev2017conditional}. We accomplish the integration of the missing viewpoint information according to the Weighted Fusion approach that finds a workable balance between the contribution of the original sequences to the final embedding and the contribution of the synthetic GAN-generated images.


As to the sequences pose-based alignment aspect, having pose-misaligned frames causes incoherent sequences matching, which results in sub-optimal performance. To address this, we propose the Weighted-Pose Regulation method based on explicitly aligning the sequence pairs based on pose/viewpoint. The two techniques, WF and WPR, leverage the integration with three state-of-the-art deep frameworks: the light-weight residual learning-based CNN model in \cite{wojke2018deep} (our baseline) for frame-level feature extraction; the key-points detector \cite{cao2016realtime} for detecting the human joints, also used in \cite{ma2017pose,siarohin2017deformable}; the deformable GAN model in \cite{siarohin2017deformable} for its ability to learn the global human body transformations across cameras. We design a balanced combination of these separate modules into a unified video-based person re-id framework and study its overall and ablative performance.

\begin{figure}
	\begin{minipage}{\columnwidth}
    \centering
        \includegraphics[width=83mm]{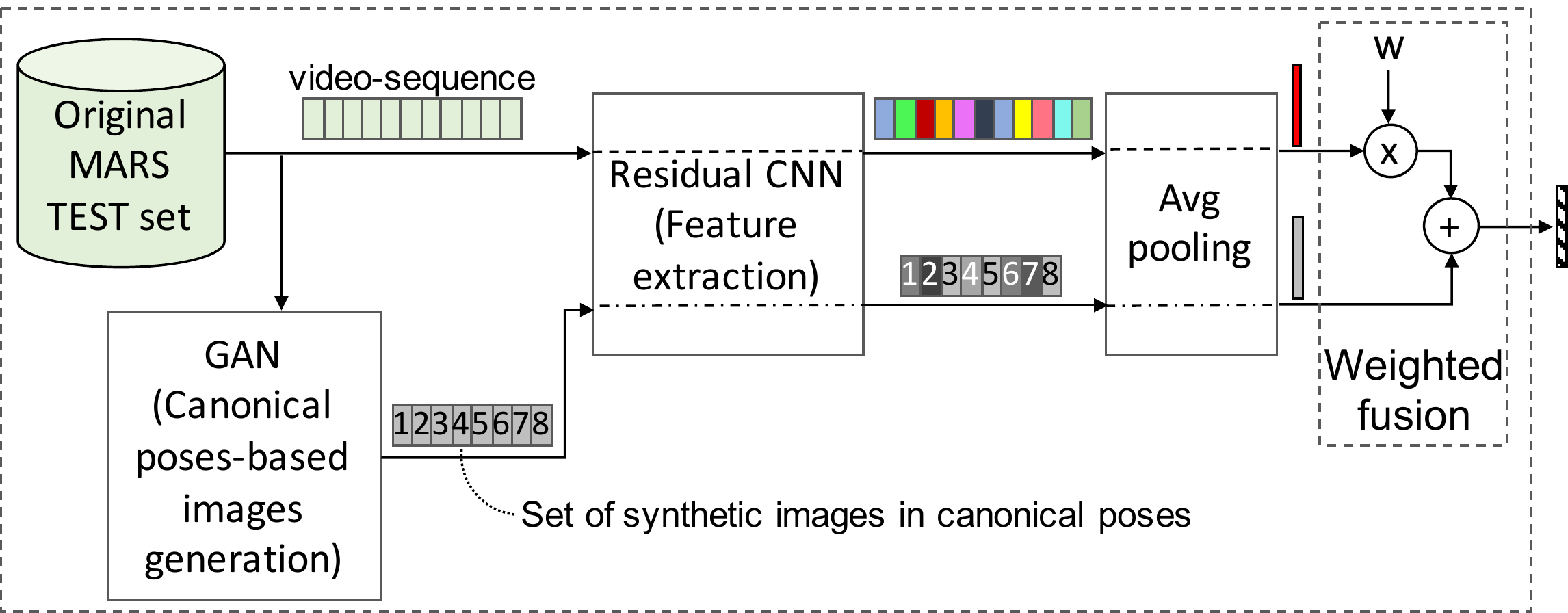}
        
		\caption{Illustration of the Weighted Fusion strategy. Best viewed in color.}
		\label{fig:WF_method}
        \end{minipage}
\end{figure}

\subsection{Weighted Fusion (WF)}
\label{sec:WF_par}
The WF scheme (Figure \ref{fig:WF_method}) is based on aggregating (by average pooling) the frame-level features of a real sequence separately from the frame-level features of the corresponding generated sequence of canonical poses, in order to exploit the complementarity of their contributions. Finally,  the two embeddings are fused together into a more discriminative representation. Using canonical poses, similarly to \cite{qian2017pose}, allows reducing the continuous variability of the real observable poses to a discretized finite pose-space for affordable pose-to-pose matching. Differently from all other papers, we operate directly on the testing set images without fine-tuning the network weights on the generated sequences in canonical form and investigate the two contributions balance. 

\begin{figure}
	\begin{minipage}{\columnwidth}
		\centering
		\includegraphics[width=80mm]{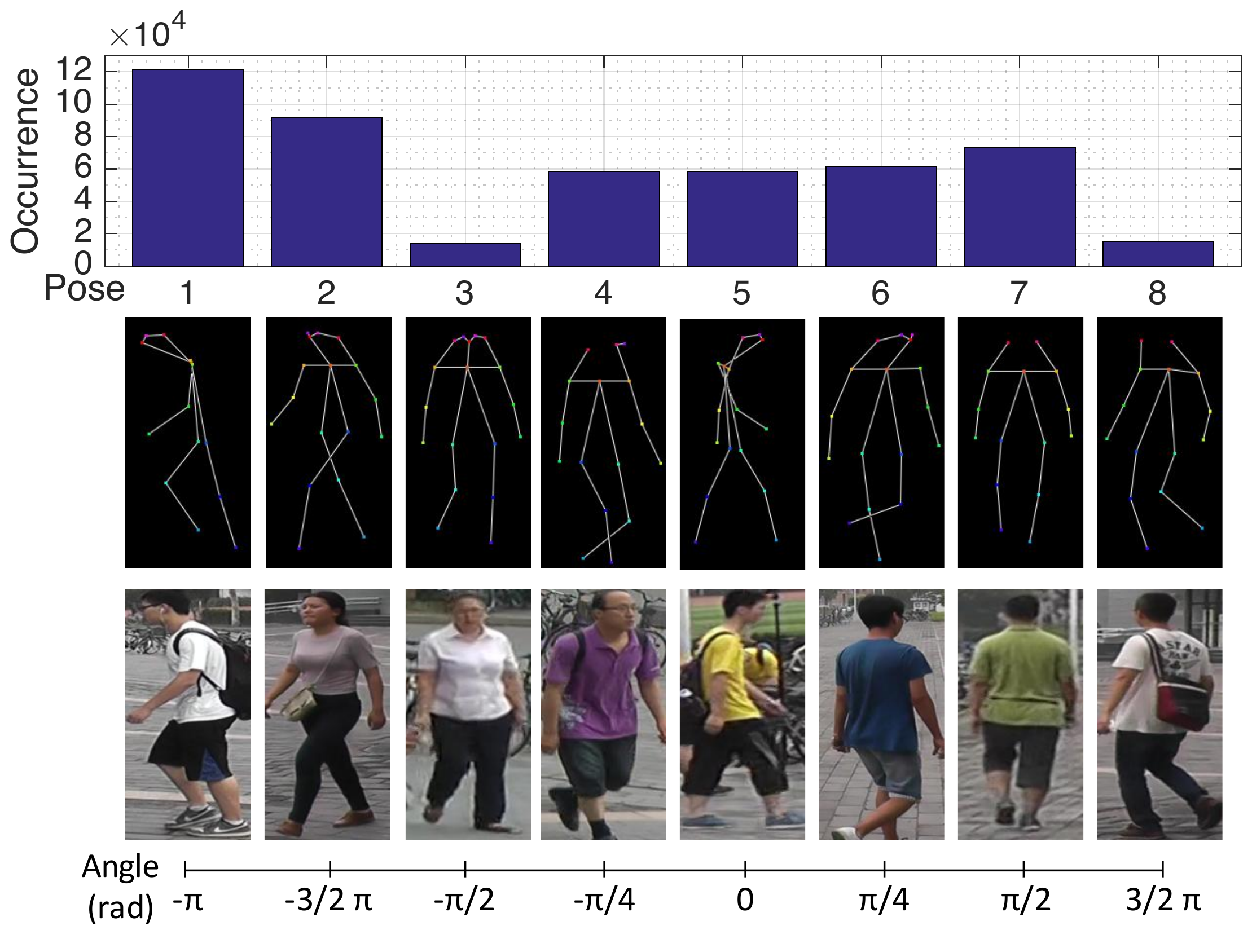}
		\caption{Illustration of the canonical poses on the MARS dataset. Top: Canonical poses distribution across the MARS dataset. Middle: The 8 canonical pose-images $\mathcal{P}({\bf \tilde{x}}_{c_j})$ that summarize the key-points information of the correspondent 8 manually selected images ${\bf \tilde{x}}_{c_j}$ from MARS (bottom). Best viewed in color.}
		\label{fig:canonical}
	\end{minipage}
\end{figure}

\begin{figure}
	\begin{minipage}{\columnwidth}
		\centering
		\includegraphics[width=73mm]{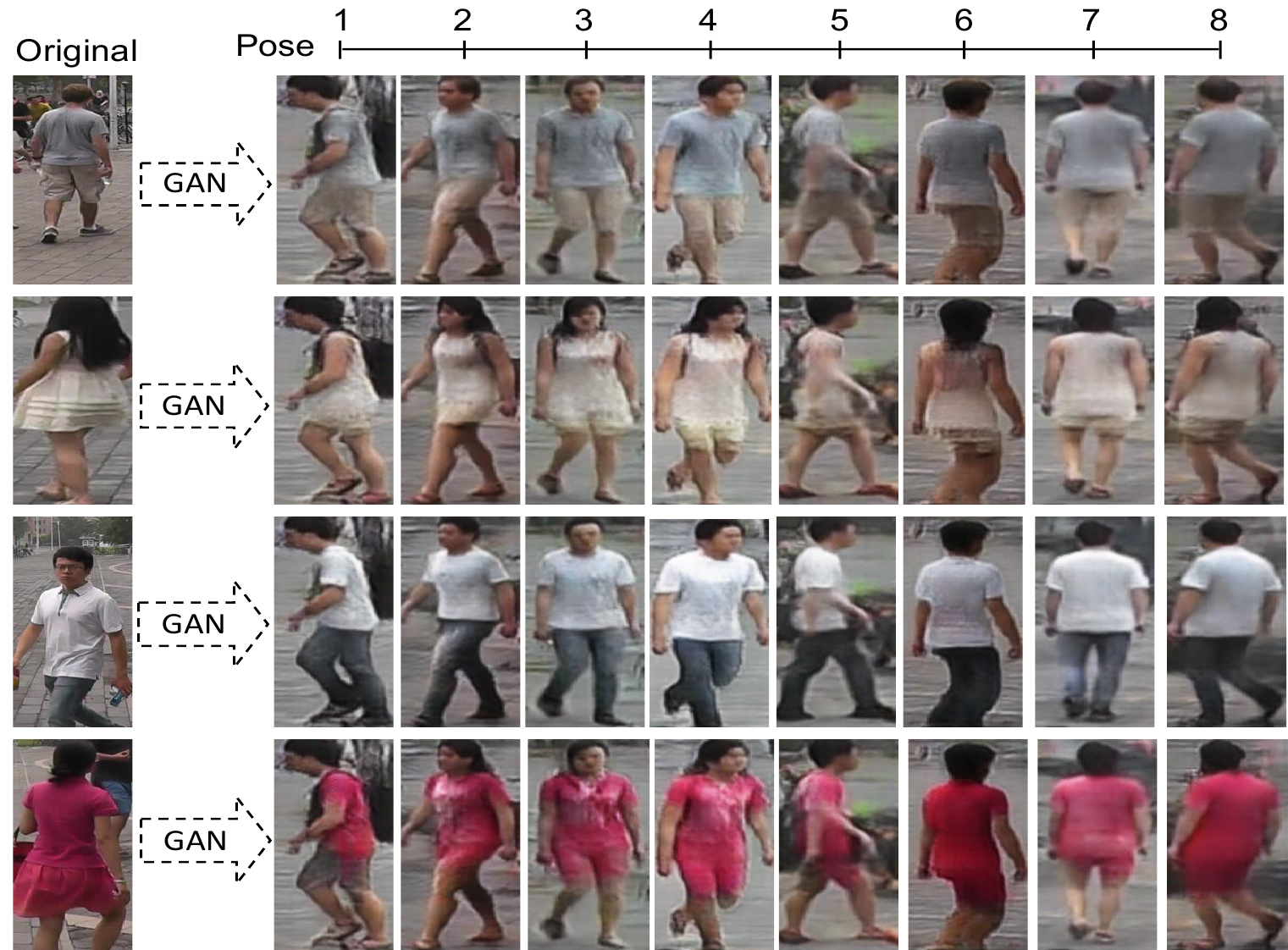}
		\caption{Translation of four MARS identities into sequences of GAN-generated images in canonical form (From the top: IDs = \{262, 1, 507, 1205\}). Best viewed in color.}
		\label{fig:generated}
	\end{minipage}
\end{figure}

Given a video-based training dataset $D_{Tr}=\{\tilde{{\bf x}}_i\}$ and a testing set $D_{Ts}=\{{\bf x}_i^{(s)}\}$, with ${\bf x}_i^{(s)} \in \mathbb{R}^d$ denoting the i-th image of the s-th video sequence, we firstly select from $D_{Tr}$ a set of M images of pedestrians $C=\{{\bf \tilde{x}}_{c_j}\}_{j=1}^{M}$, $c_j\in \{1,...,|D_{Tr}|\}$ representative of M significant canonical viewpoints in which the pose space can be quantized (Figure \ref{fig:canonical}, bottom row). The different viewpoints are intended as angular rotations around the imaginary longitudinal axis drawable across the human body. Secondly, for each ${\bf \tilde{x}}_{c_j} \in C$, the key-points detector proposed in \cite{cao2016realtime} is used to extract a 2D coordinates vector of k joints and mapping it to a pose-image $\mathcal{P}({\bf \tilde{x}}_{c_j})$ that summarizes the key-points information and that we refer to as canonical image (Figure \ref{fig:canonical}, middle row). Thirdly, the heat maps $\mathcal{H}(\mathcal{P}({\bf \tilde{x}}_{c_j}))$ of the canonical poses, representing the pose-conditioning input to the GAN, are calculated. Fourthly, for each testing sequence, one image ${\bf x}_i^{(s)}$ representative of the pedestrian identity is randomly drawn and used to generate a set of M synthetic images $\{{\bf \hat{x}}_{i,j}^{(s)}\}_{j=1}^{M}$ with the same identity as ${\bf x}_i^{(s)}$ but represented under the M canonical poses (Figure \ref{fig:generated}). For the generation of these images three inputs to the GAN are required: ${\bf x}_i^{(s)}$ with its associated pose-image $\mathcal{H}(\mathcal{P}({\bf x}_i^{(s)}))$ (as detailed in \cite{siarohin2017deformable}) and the heat maps of the canonical poses. This generation process creates a new synthetic testset $\hat{D}_{T_s}$ corresponding, sequence-by-sequence, to the original one $D_{T_s}$, which is exploitable to extract video-level features ${\bf h}^{(s)}$ more robust from the viewpoint invariance perspective. The features extracted from the s-th test sequence can be mathematically expressed as in Equation \ref{eq:enh_feat}:

\begin{equation}
\label{eq:enh_feat}
\begin{split}
{\bf h}^{(s)} = w \cdot [\frac{1}{L}\bigoplus_{i=1}^{L}f({\bf x}_i^{(s)})] \oplus [\frac{1}{M}\bigoplus_{j=1}^{M}f({\bf \hat{x}}_{i,j}^{(s)})]
\end{split}
\end{equation}

\noindent where $\bigoplus$ indicates the Kronecker element-wise sum operation of multiple matrices (feature maps); $\oplus$ denotes the element-wise sum of two matrices; $w$ is the weighting parameter of the WF method; $L$ is the length of the original sequence that ${\bf x}_i$ belongs to; $f(\cdot)$ represents the feature extraction function of the CNN learned on the training set. 

The synthetic image ${\bf \hat{x}}_i$ can be represented as in Equation \ref{eq:new}:

\begin{equation}
\label{eq:new}
\begin{split}
{\bf \hat{x}}_i=G({\bf z},{\bf x}_i,\mathcal{H}_i, \tilde{\mathcal{H}}_j)
\end{split}
\end{equation}

\noindent where $G$ is the generator of the GAN; ${\bf z}$ is a noise vector that is implicitly incorporated by
the network dropout; $\mathcal{H}_i\equiv\mathcal{H}(\mathcal{P}({\bf x}_i))$ and $\tilde{\mathcal{H}}_j\equiv\mathcal{H}(\mathcal{P}({\bf \tilde{x}}_j))$ are the heat maps of a generic image and of one canonical pose image belonging to $C$, respectively. The generator optimization is driven by two losses, the standard conditional adversarial loss $\mathcal{L}_{GAN}(G,D)$ and the nearest-neighbour loss $\mathcal{L}_{NN}(G)$ defined in \cite{siarohin2017deformable}, as in Equation \ref{eq:linCombLoss}:

\begin{figure}
	\begin{minipage}{\columnwidth}
		\centering
		\includegraphics[width=84mm]{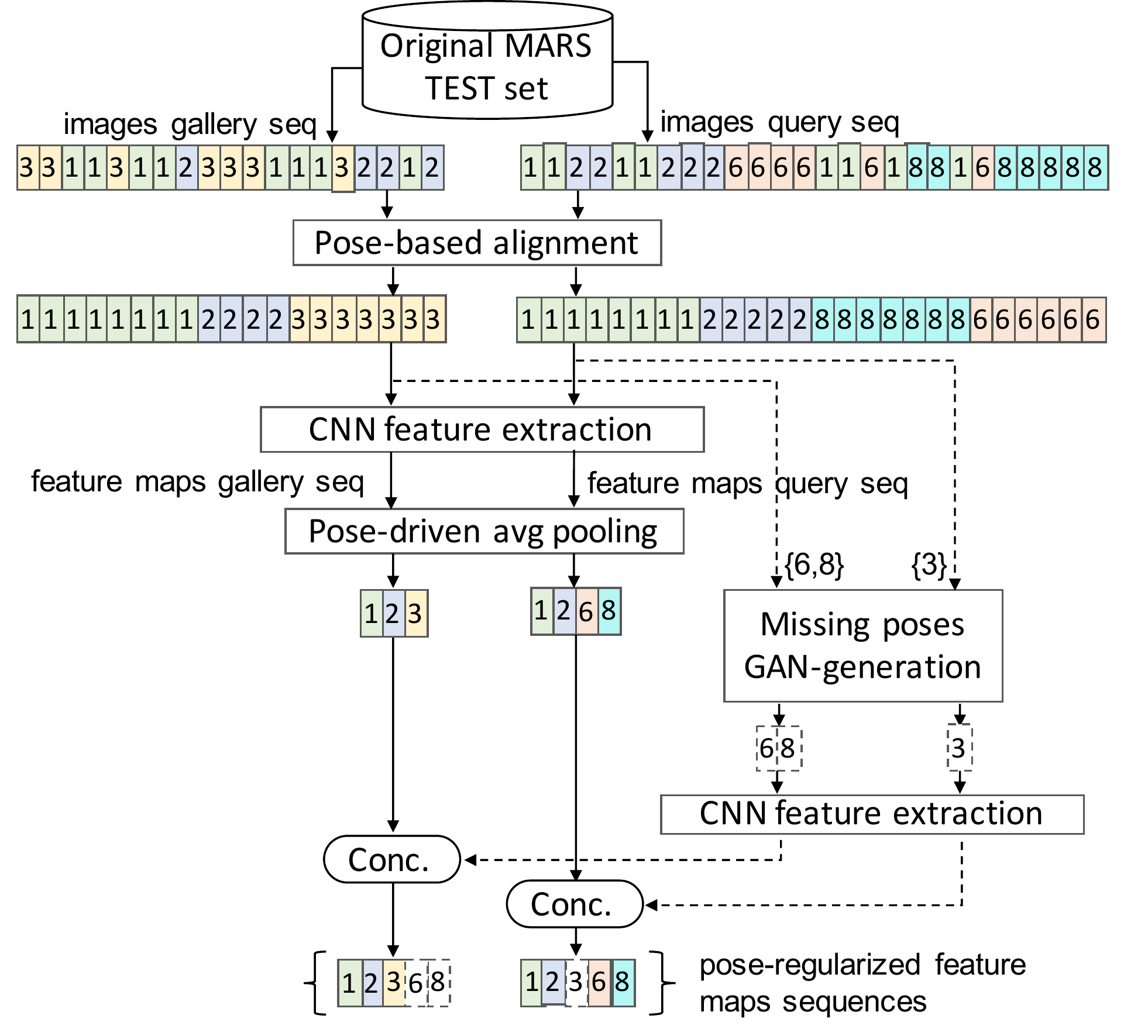}
		\caption{Pose-based sequence pair normalization block.}
		\label{fig:seq_pair_norm}
	\end{minipage}
\end{figure}

\begin{equation}
\label{eq:linCombLoss}
\begin{split}
G^{*}= \arg \min\limits_{G}\max\limits_{D}\mathcal{L}_{GAN}(G,D) +\lambda \mathcal{L}_{NN}(G)
\end{split}
\end{equation}

\noindent where $D$ is the GAN discriminator and $\lambda$ the coefficient of the linear combination. It is noteworthy to underline that the weighting parameter $w$ (explored throughout its space of variation) plays the critical role of controlling the level of noise introduced into the deep representation by the generated images, in order to make effective fusion.

\begin{figure}
	\begin{minipage}{\columnwidth}
		\centering
		\includegraphics[width=40mm]{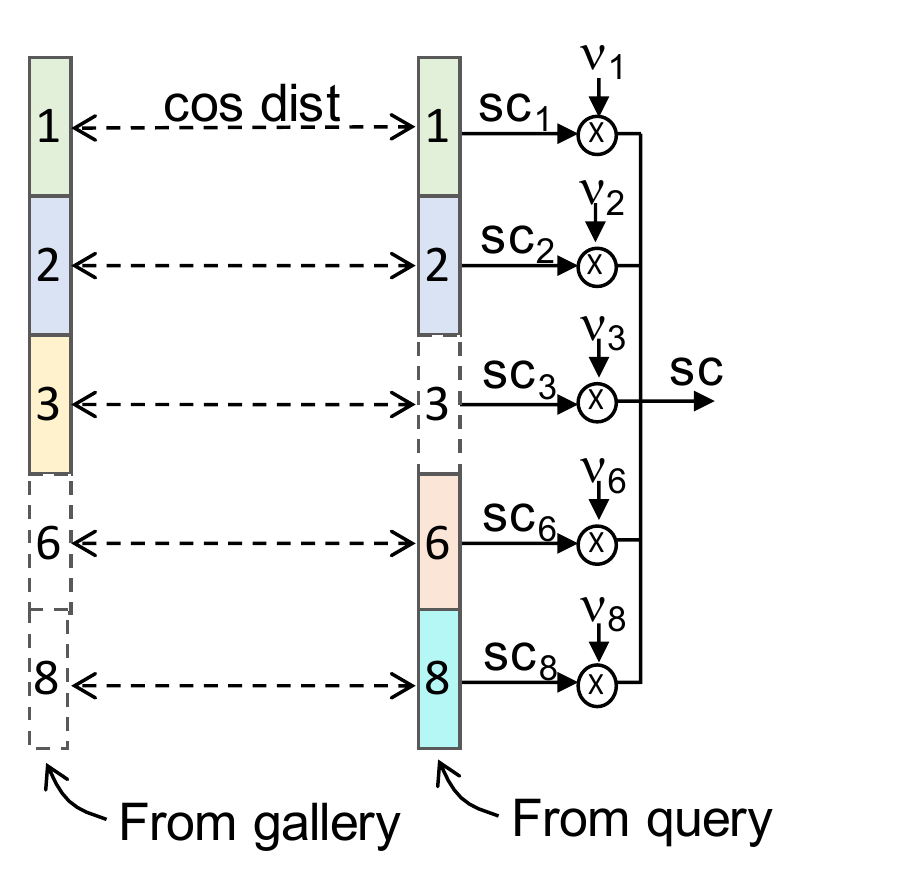}
		\caption{Matching strategy. One pair of pose-aligned sequences of feature maps are matched pose by pose by the cosine distance.}
		\label{fig:score_scheme}
	\end{minipage}
\end{figure}

\subsection{Weighted Pose-Regulation (WPR)}
The sequence pair alignment method based on the viewpoint factor represents an independent source of performance improvement. We propose a scheme articulated into two blocks: one block performs viewpoint-based sequence pair normalization (Figure \ref{fig:seq_pair_norm}), and another deals with the actual feature vectors matching (Figure \ref{fig:score_scheme}). As to the sequence normalization block, given a pair of tracklets $X_{s_1}$ and $X_{s_2}$ of arbitrary length to be matched, each of them is reassembled into up to M view-point specific sub-sequences $X_s=[X_s^{(p_1)}, ..., X_s^{(p_j)}],$ $1\leqslant j\leqslant M $, grouping images that share the same canonical pose $p_j\equiv \mathcal{P}({\bf \tilde{x}}_{c_j})$. With regards to the canonical poses introduced in Section \ref{sec:WF_par}, one image ${{\bf x}_i^{(s)}}$ is considered to have the j-th canonical pose if the Euclidean distance of its extracted key-points vector from it is the smallest, that is: $|\mathcal{P}({{\bf x}_i^{(s)}})-\mathcal{P}({{\bf \hat{x}}_{c_j}^{(s)}})|\leqslant|\mathcal{P}({{\bf x}_i^{(s)}})-\mathcal{P}({{\bf \hat{x}}_{c_l}^{(s)}})|$, $\forall j\neq l$, $j,l\in \{1,..., M\}$, with $|\cdot|$ denoting the Euclidean distance. By performing frame-level feature extraction $f(\cdot)$ and average pooling $avg(\cdot)$ separately on each single pose-specific sub-sequence $X_s^{(p_j)}$, a video-based representation ${\bf f}^{(s)}\equiv[avg(f(X_s^{(p_1)})), ..., avg(f(X_s^{(p_j)}))]\equiv[{\bf f}^{(s,p_1)},..., {\bf f}^{(s,p_j)}]$ is produced as the concatenation of view-specific contributions ${\bf f}^{(s,p_j)}$. At this stage, the video-level multi-pose aggregated representations ${\bf f}^{(s_1)}$ and ${\bf f}^{(s_2)}$ of the two sequences $X_{s_1}$ and $X_{s_2}$ are not yet pose-aligned, because each one includes a different sub-set of canonical poses $R^{(s)}\equiv\{p_j\}_{j \in \{1,..., M\}}$. In order to align them, they both must include, respectively in ${\bf f}^{(s_1)}$ and ${\bf f}^{(s_2)}$, one video-feature map for each of the canonical poses in $R^{(s_1)} \cup R^{(s_2)}$. The missing ones are GAN-generated and concatenated. Denoting by $Q^{(s)}\equiv\{p_m\}_{m \in \{1,..., M\}}$ the set of the missing poses in ${\bf f}^{(s)}$, with $Q^{(s)} \cap R^{(s)}=\emptyset$ and $Q^{(s)} \cup R^{(s)}=C$,		 we can express the final pose-normalized sequence representation as in Equation \ref{eq:WPR_eq}

\begin{equation}
\label{eq:WPR_eq}
\begin{split}
\overline{{\bf h}}^{(s)}=cat_{sort}([ {\bf f}^{(s)},..., \{{\bf f}^{(s,p_m)}\}_{m \in Q^{(s)}}])
\end{split}
\end{equation}

\noindent where $cat_{sort}(\cdot)$ is a function that performs vertical concatenation of the ${\bf f}^{(s,p_j)}$ by sorting all terms according to the increasing pose index $j=1,...,M$. With regards to the matching/ranking block, we design an ad-hoc strategy for matching pairs of pose-normalized feature map sequences $\overline{{\bf h}}^{(s_1)}$, $\overline{{\bf h}}^{(s_2)}$, based on performing separate sub-matching of pose-specific feature maps $\{{\bf f}^{(s,p_m)}\}$ according to how illustrated in Figure \ref{fig:score_scheme}, where the weight parameter $\nu_j$ represents the averaged value of the frequencies with which the canonical pose $p_j$ occurs in the two matching sequences.



\begin{figure}
	\begin{center}
		\includegraphics[width=75mm]{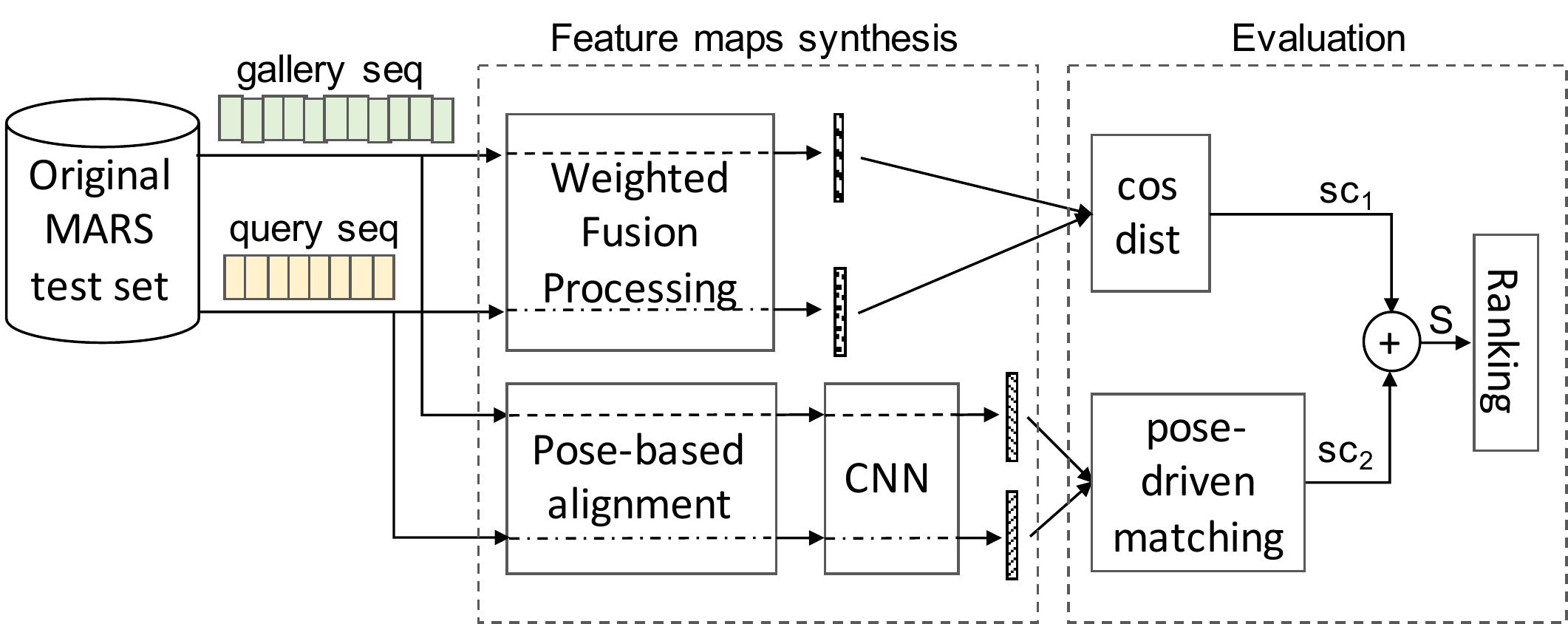}
        \end{center}
		\caption{Overall system architecture. The pose-driven matching block is shown in Figure \ref{fig:score_scheme}, the Pose-based alignment block in Figure \ref{fig:seq_pair_norm} and the weighted fusion processing in Figure \ref{fig:WF_method}.}
		\label{fig:system_combined}
\end{figure}

\subsection{Combining WF and WPR}
Due to the conceptual complementarity of the two methods, we can combine them together, shown in Figure \ref{fig:system_combined}, to benefit from both additive contributions at the same time. The fusion happens at the score level: the two score vector produced by WF and WPR (using the cosine distance) are summed up element-wise and ranked by decreasing overall score. With this approach, the final ranking list accounts for both pose information completeness and pose alignment, promoting progress in the ranking for those feature matchings that would get penalized by the single WF score calculation because of their frame-based pose misalignment.

\section{Experiment}

\subsection{Database}
\label{database}
MARS (Motion Analysis and Re-identification Set) \cite{zheng2016mars} is one of the largest video re-id datasets currently available and represents an extension of Market-1501 \cite{zheng2015scalable}. It contains $1,261$ IDs (631 belonging to the training set and 630 to the testing set) captured by at least two of the $6$ cameras deployed. The $20,715$ video-sequences of MARS are automatically generated by the Deformable Part Model pedestrian detector \cite{felzenszwalb2010object} and the GMMCP tracker \cite{dehghan2015gmmcp}. MARS reproduces the challenges of a real-world scenario, due to the presence of $3,248$ distractors and of many partial and total occlusions that make its tracklets quite noisy.

DukeMTMC-VideoReID \cite{wu2018exploit} is a large-scale video-based re-identification dataset, created from the multi-target multi-camera tracking dataset DukeMTMC \cite{ristani2016performance}, that counts $1,404$ valid IDs captured by up to $8$ disjoint static cameras plus $408$ IDs which are distractors appearing in only one camera. The pedestrians are randomly split into a training set and a testing set, each one with $702$ IDs. The training set is structured in $2,196$ tracklets made up of 369,656 frames, while the testing set includes $2,636$ tracklets formed by $445,764$ frames.

\begin{figure}
	\begin{center}
        \includegraphics[width=80mm]{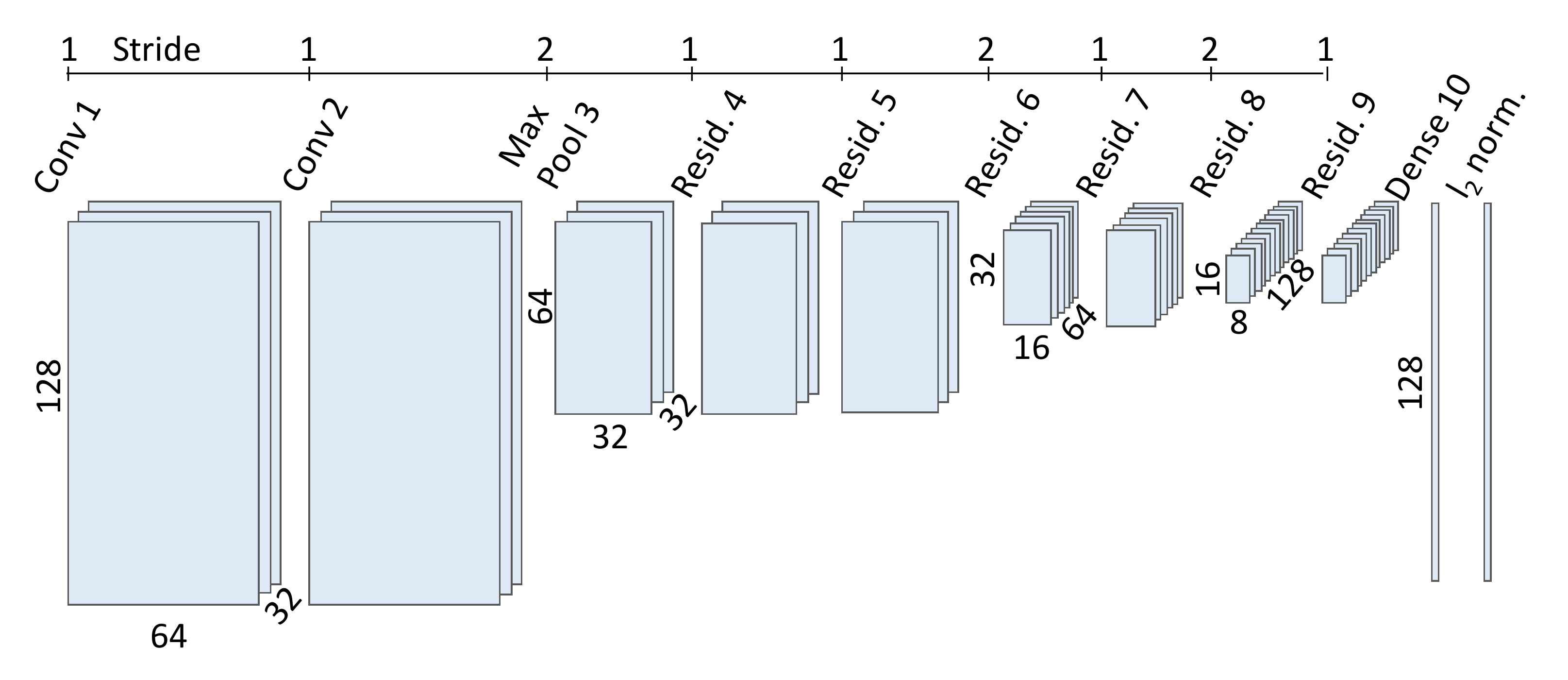}
        \end{center}
		\caption{Residual learning-based CNN architecture \cite{wojke2018deep} used for feature extraction. The patch size for the convolutional, pooling and residual layers is $3\times 3$.}
		\label{fig:cnn}
\end{figure}

\subsection{Evaluation Metrics and Protocols}
\label{protocols}
The evaluation protocol of MARS and DukeMTMC-VideoReID datasets are borrowed from the Market1501 dataset \cite{zheng2015scalable} with the difference that, for the video-based datasets, probes and queries are tracklets instead of still images. For each identity, one query tracklet representing that identity is randomly selected and tested against the gallery sequences which are finally ranked based on their cosine similarity to the query. The overall performance is calculated by averaging across all identities. The gallery sets of the probes include all the sequences belonging to camera views different to the probe, including distractors. The metrics that we use for evaluating our person re-id algorithms against both datasets are the Cumulated Matching Characteristics (CMC) curve and the mean-Average Precision (mAP) in order to account jointly for precision and recall. Furthermore, in order to measure the relative improvement of re-id between camera pairs, we build the mAP camera-pairs confusion matrix, shown in Figure \ref{fig:mAP_cnf_mtr}, for the WF+WPR method on MARS against the DCML cosine-softmax baseline \cite{wojke2018deep}. For a fixed pair of cameras $(X,Y)$, where $X$ is the probe camera and $Y$ represents the test camera, the corresponding value in the confusion matrix is calculated by limiting the positive samples (tracklets) to only those ones viewed under camera $Y$ and averaging for all the probes viewed under the same camera $X$.

\subsection{Implementation Details}
\label{implementation}
\paragraph{Feature extraction CNN.} The CNN architecture that we use for feature extraction is the one proposed in \cite{wojke2018deep}, shown in Figure \ref{fig:cnn}, based on the residual learning scheme \cite{He2016} and principles \cite{zagoruyko2016wide}. It is a quite shallow network made up of only 15 layers in order to be suitable for the task of people tracking which requires fast features extraction capabilities. Dropout \cite{srivastava2014dropout} and batch normalization \cite{ioffe2015batch} are applied between layers for regularization purposes. The input is represented by RGB images rescaled to $128 \times 64$. The cosine-softmax classifier replaces the standard softmax classifier in order to get more compact classes in the feature space representation \cite{wojke2018deep}. The CNN is trained directly on the target re-id dataset with a batch size of $128$ images. For training the network for the identity classification task, $631$ classes for MARS and $702$ for DukeMTMC-VideoReID are set and the output of the $l_2$ normalization layer (i.e., a 128 elements vector) is used as feature embedding for the following matching step.

The training is regulated by a learning rate $\eta=10^{-3}$ and the network was regularized by a weight decay of $10^{-1}$ and a dropout with probability $40\%$. The model checkpoint at 46461 iterations is selected as the best one.

\begin{table}
	\begin{center}
      \caption{State-of-the-art (\%) on the MARS. $^{(a)}$ indicates a training/architectural advantage due to the use of Imagenet pre-training. $^{(b)}$ denotes semi-unsupervised methods.}
      \label{table:state_of_the_art}
      \scalebox{0.90}{
      \begin{tabular}{l|ll}
        \textbf{Method} & \textbf{Rank 1} & \textbf{mAP} \\ \hline
        CAR + Video \cite{zhang2017learning} & 55.5 & \hspace{3mm}- \\
        ETAP Net \cite{wu2018exploit} $^{(a,b)}$ & 62.67 & 42.45 \\
        IDE (C) + XQDA \cite{zheng2016mars} $^{(a)}$ & 65.30 & 47.60 \\
        IDE (R) + ML \cite{zhong2017identification} & 70.51 & 55.12 \\
        CaffeNet \cite{zhou2017see} & 70.60 & 50.70 \\
        MSCAN \cite{li2017learning} & 71.77 & 56.06 \\
        IDE (R) + ML \cite{hermans2017defense}& 72.42 & 57.42 \\
        DCML (baseline) \cite{wojke2018deep}& 72.93 & 56.88 \\
        P-QAN \cite{liu2017quality} & 73.73 & 51.70 \\
        \textbf{Ours} & \textbf{75.76} & \textbf{60.57} \\ \hline
      \end{tabular}}
    \end{center}
\end{table}

\paragraph{GAN model.} 
For generating synthetic images, we borrow the deformable GAN presented in \cite{siarohin2017deformable} because it bypasses the commonly required two stages training that is replaced by a single stage end-to-end training. The generator and the discriminator are trained on the Market-1501 dataset for $90,000$ iterations, with batch size equal to 4 and warping skip layers of type ``mask" that allows applying the warping only to the foreground target removing the background, like in \cite{ma2017pose}. The optimization is performed by the Adam optimizer with learning rate $\alpha=2*10^{-4}$, exponential decay rate for the first moment $\beta_1=0.5$ and for the second moment $\beta_2=0.999$. We train the GAN on Market-1501 instead of directly on MARS because the latter represents a more noisy video-based extension of the Market-1501.

\paragraph{Pose estimation model.} 
For pose estimation we use the real-time multi-person model (Human Pose Estimator, HPE) proposed in \cite{cao2016realtime} and also used in \cite{siarohin2017deformable, ma2017pose} that is based on a non-parametric representation of 18 landmarks corresponding to the human body joints locations.

\begin{table}
	\begin{center}
      \caption{State-of-the-art (\%) on the DukeMTMC-VideoReID dataset. $^{(a)}$ denotes training/architectural advantage due to the use of Imagenet pre-training and a deeper network. $^{(b)}$ denotes semi-unsupervised methods.}
      \label{table:duke_state_of_the_art}
      \scalebox{0.90}{
      \begin{tabular}{l|ll}
        \textbf{Method} & \textbf{Rank 1} & \textbf{mAP} \\ \hline
        DGM+IDE \cite{ye2017dynamic} $^{(a)}$ & 42.36 & 33.62 \\
        Stepwise \cite{liu2017stepwise} $^{(a)}$ & 56.26 & 46.76 \\
        ETAP Net \cite{wu2018exploit} $^{(a,b)}$ & 72.79 & 63.23 \\
        \textbf{Ours} & \textbf{82.22} & \textbf{78.76} \\ \hline
      \end{tabular}}
    \end{center}
\end{table}

\subsection{Comparison with State-of-the-art Methods}

Our method, relying on a light-weight CNN, outperforms most of the state-of-the-art techniques in video-based person re-id on MARS (Table \ref{table:state_of_the_art}) and on DukeMTMC-VideoReID (Table \ref{table:duke_state_of_the_art}). With regards to the MARS dataset, we achieve a rank 1 accuracy of $75.76\%$ and a mAP of $60.57\%$ with an improvement respectively of $+2\%$ and $+8.9\%$ with respect to the second best method \cite{liu2017quality}. For the sake of completeness, we should clarify that we exclude from the comparison a few methods that are not directly comparable with ours because they use more powerful learning strategies with more data, or deeper and more complex architectures (Siamese): one examples is  \cite{hermans2017defense}, excluded from Table \ref{table:state_of_the_art} because it employs a triplet network, although, despite that, our results are still higher than those of its LuNet network ($rank_1= 75.56\%$ and $mAP=60.48\%$). Another examples is \cite{li2018diversity} that uses ResNet50 pre-trained on 6 re-id datasets and fine-tuned on three additional ones with overall four stages training.

Also on the DukeMTMC-VideoReID dataset, both WF and WPR succeed in boosting the performance of our baseline, reaching the state-of-the-art on this very new dataset (Table \ref{table:duke_state_of_the_art}). We achieve an improvement of $+9.4\%$ and $+15.5\%$ respectively for rank 1 accuracy and mAP, with regards to the second best performing method, ETAP Net \cite{wu2018exploit}. For a fair comparison, it should be pointed out that \cite{wu2018exploit} is a semi-unsupervised technique while ours is a supervised one. By the way, ETAP Net provides also a supervised upper bound ($83.62\%$ and $78.34\%$ for rank 1 accuracy and mAP respectively) which is achieved with a much deeper network, ResNet50, pre-trained on the Imagenet dataset. Despite our method relies on a network with a depth of around $1/3$ of ResNet50 and is trained directly on the target train set, compared to the ETAP Net supervised upper bound, we reduce the architecture structure disadvantage and achieve competing results ($-1.4\%$ $+0.4\%$ for rank 1 accuracy and mAP respectively). Our results show that the features extracted from GAN-generated images with a model learned only on the real images can be used for identity discrimination at testing time other than for network regularization at training time as in \cite{zheng2017unlabeled}, maintaining the identity label of their original conditioning images. This makes our approach easy to be applied to different models for getting a boost in performance with no need of retraining. It is worth noting that adding the M synthetic canonical poses to each identity sub-sequence in the \textit{train set} for data augmentation, does not improve the performance. Furthermore, it negatively affects the re-id, even for small values of the augmentation factor. We reckon that this is due to the high level of noise introduced into the learning process by the less discriminative synthetic images which make the learned video-level features more ambiguous. Exploiting GAN-generated images for data-augmentation in person re-id has been done in \cite{zhong2018camera} for camera style adaptation. A substantial difference between the camera style transfer case and ours, though, is that while in \cite{zhong2018camera} the synthetic images are conditioned on the camera style and thus preserve the real content of the original conditioning images, in our work the conditioning is done on pose which is a more difficult characteristic to reproduce while preserving the identity, because it involves an image content change.

\begin{table}
	\begin{center}
      \caption{Ablation study on MARS w.r.t WF and WPR (\%).}
      \label{table:ablation}
      \scalebox{0.90}{
      \begin{tabular}{l|ll}
      \textbf{Method} & \textbf{Rank 1} & \textbf{mAP} \\ \hline
      DCML$^{(*)}$ (baseline) & 72.57 & 56.57 \\ \hline
      WF (ours) & 74.19 (+1.6) & 59.48 (+2.9)\\
      \textbf{WF + WPR (ours)} & \textbf{75.76  (+3.2)} & \textbf{60.57 (+4)} \\ \hline
      \end{tabular}}
	\end{center}
\end{table}

\begin{table}
	\begin{center}
      \caption{Ablation study on DukeMTMC-VideoReID (\%).}
      \label{table:duke_ablation}
      \scalebox{0.90}{
      \begin{tabular}{l|ll}
      \textbf{Method} & \textbf{Rank 1} & \textbf{mAP} \\ \hline
      DCML$^{(*)}$ (baseline) & 79.20 & 75.25  \\ \hline
      WF (ours) & 80.84 (+1.6) & 77.56 (+2.3)\\
      \textbf{WF + WPR (ours)} & \textbf{82.22  (+3)} & \textbf{78.76 (+3.5)} \\ \hline
      \end{tabular}}
	\end{center}
\end{table}

\subsection{Ablation Analysis}
\label{sec:ablation_sec}

On DukeMTMC-VideoReID, both WF and WPR improve the baseline ($+79.20\%$ and $75.25\%$ respectively for rank 1 accuracy and mAP) that simply performs the average pooling of the frame-level features. Table \ref{table:duke_ablation} shows that $53\%$ of the rank 1 accuracy improvement ($66\%$ for the mAP) is due to the WF method, while the remaining $+47\%$ ($34\%$ for mAP) to the WPR viewpoint-based alignment technique, which highlights the effectiveness of our two schemes and demonstrates that they are complementary to some extent.

\begin{figure}
	\begin{minipage}{\columnwidth}
		\centering
		\includegraphics[width=80mm]{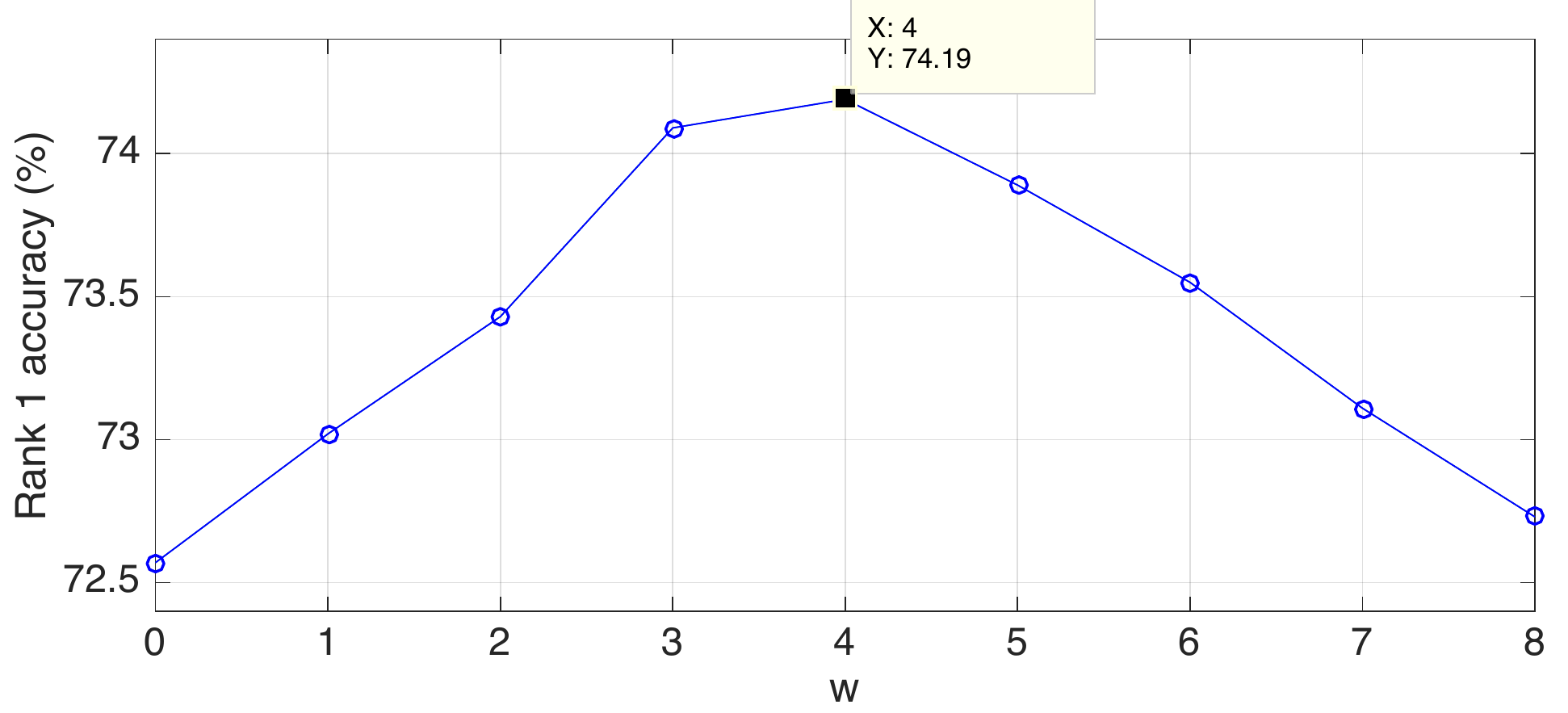}
		\caption{Rank 1 accuracy on MARS. Baseline @$w=0$.}
		\label{fig:rank1_mars}
	\end{minipage}
\end{figure}

\begin{figure}
	\begin{center}
		\includegraphics[width=84mm]{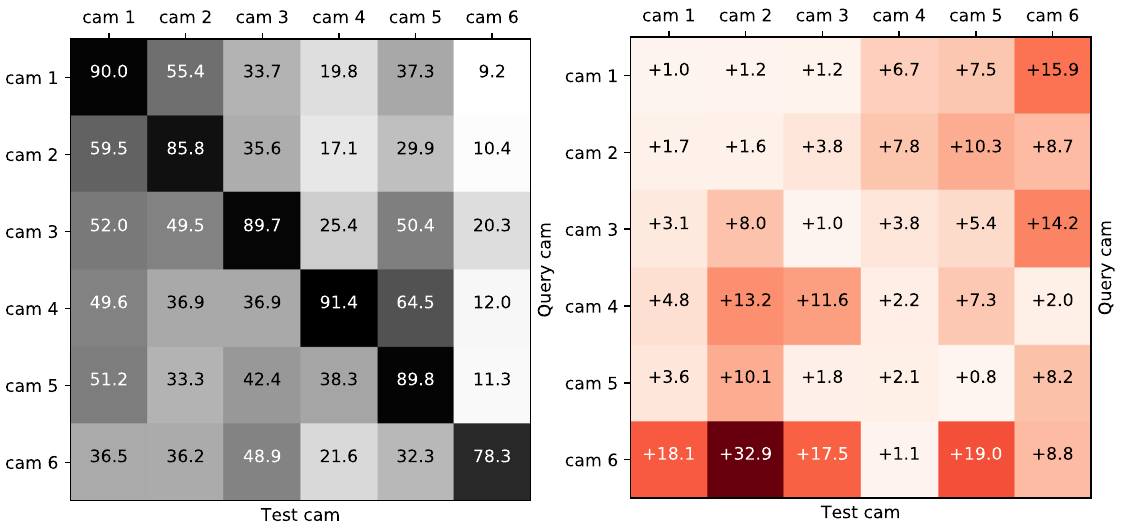}  
		\caption{Left: mAP confusion matrix (\%) for DCML \cite{wojke2018deep}. Right: mAP relative confusion matrix for WF+WPR w.r.t. DCML.}
		\label{fig:mAP_cnf_mtr}
	\end{center}
\end{figure}


With regards to the baseline \cite{wojke2018deep}, on MARS, our PDSR technique allows an overall improvement of $+3.2\%$ and $+4\%$ respectively for the rank-1 accuracy and the mAP Table \ref{table:ablation}. This improvement is due in part to the weighted fusion strategy WF ($+1.6\%$ and $+2.9\%$ over the baseline, respectively, for rank 1 accuracy and mAP) and for the remaining part to the WPR technique ($+1.6\%$ and $+1.1\%$ for rank-1 accuracy and mAP). With regards to WF, we report in Figure \ref{fig:rank1_mars} the curve describing how the rank-1 accuracy varies over the weighting parameter space of the linear combination. The top value, corresponding to $w=4$, defines the WF performance. Evidence of the benefit of our method to mitigate the effects of the inter-camera viewpoint problem when matching pedestrians tracklets comes from the mAP confusion matrix in Figure \ref{fig:mAP_cnf_mtr} for it shows that the highest relative improvements to the re-id performance happen for the cross-camera matchings compared to the intra-camera re-id cases (confusion matrix main diagonal).

\section{Conclusions}
In this paper, with regards to the video-based person re-id task, we addressed the problem of how to account effectively for the pose/viewpoint information in the pose-incomplete and noisy video-sequences matching process at the testing time. We formulated two separate approaches, WF and WPR, that work also jointly and rely on the definition of canonical poses, weight-controlled fusion, generated canonical poses sequences and viewpoint-based sequences alignment. Combined, our techniques achieve state-of-the-art performance on MARS and DukeMTMC-VideoReID.


\section*{Acknowledgement}
This work was supported in part by the Engineering and Physical Sciences Research Council (EPSRC), Grant number EP/K014277/1, in part by the MOD University Defence Research Collaboration (UDRC) in Signal Processing and in part by Roke Manor Research.


{\small
\bibliographystyle{ieee}
\bibliography{reid.bib}
}

\end{document}